\documentclass{article} %
\usepackage{iclr2027_conference,times}

\usepackage{amsmath,amsfonts,bm}

\def\eqref#1{equation~\ref{#1}}

\def\1{\bm{1}}

\DeclareMathAlphabet{\mathsfit}{\encodingdefault}{\sfdefault}{m}{sl}
\SetMathAlphabet{\mathsfit}{bold}{\encodingdefault}{\sfdefault}{bx}{n}

\usepackage{graphicx}
\usepackage{subcaption}
\usepackage{wrapfig}
\usepackage{amsmath}
\usepackage{adjustbox} %
\usepackage{listings}

\usepackage{hyperref}
\usepackage{url}
\usepackage[utf8]{inputenc} %
\usepackage[T1]{fontenc}    %
\usepackage{hyperref}       %
\usepackage{url}            %
\usepackage{booktabs}       %
\usepackage{amsfonts}       %
\usepackage{nicefrac}       %
\usepackage{microtype}      %
\usepackage[dvipsnames]{xcolor}
\usepackage{cleveref}
\usepackage{multirow}

\title{Equivariance Breaks the Learning Rate}

\author{Andrei Manolache\thanks{Correspondence to: \texttt{andrei.manolache@ki.uni-stuttgart.de}.} \\
University of Stuttgart, Germany \\
Bitdefender, Romania \\
\And
Mathias Niepert \\
University of Stuttgart, Germany
}

\iclrfinalcopy
\begin{document}

\maketitle
\lhead{Preprint}

\begin{abstract}

	Equivariant networks are commonly trained with Adam, yet recent work reports that matrix structured optimizers such as Muon can perform better, with the reasons for these gains only partly understood.
	We identify one source of this difference inside equivariant layers.
	An equivariant layer learns one channel mixing matrix $W_l$ per degree $l$, which we call an irrep block, and shares it across the $2l+1$ components, giving the expanded map $W_l \otimes I_{2l+1}$.
	This sharing sums gradient contributions across components and can produce different update scales under SGD.
	Adam's entrywise normalization reduces sensitivity to gradient scale, but neither optimizer directly controls the effective step size of each block.
	A single learning rate can therefore produce different effective step sizes across blocks.
	Muon instead controls the effective step size by approximately equalizing the singular values of each momentum matrix.
	We normalize each irrep block update by a single scalar, preserving its singular value ratios while letting the learning rate control its size.
	We implement this with spectral normalization or a simpler root-mean-square normalization.
	We evaluate spectral normalization in a controlled $\mathrm{SO}(3)$-equivariant model with a matched non-equivariant model.
	In this setting, the step size mismatch grows with width in the equivariant model but not in the non-equivariant model.
	We evaluate both variants across molecular force prediction on the rMD17 and MD22 datasets, QM9 molecular property prediction, and charged particle dynamics.
	Across these applications, block normalization generally improves Adam and closes part of its gap to Muon.
	These results highlight an overlooked interaction between equivariant architectures and their optimizers. Studying and designing the two together may help explain and address training difficulties often attributed to equivariance itself.
\end{abstract}

\section{Introduction}

Equivariant neural networks encode known geometric symmetries directly into their architecture, providing a strong inductive bias for learning from structured data~\citep{thomas18,pmlr-v139-satorras21a,bronstein21}.
Equivariance is particularly important for physical sciences, where most models' predictions should not depend on the choice of coordinate frame, hence equivariant models have become widely used for molecular property prediction, atomistic simulation, and learned interatomic potentials~\citep{aykent2025gotennet,batzner2022nequip,batatia2022mace}.
Yet hard equivariance constraints can also make these models difficult to optimize, producing unfavorable loss geometry and sometimes underperforming less constrained alternatives at scale~\citep{petrache2023approximationgeneralization,xie2025a,brehmer2025doesequivariancematterscale}. Much recent work addresses this difficulty by modifying the architecture through approximate or relaxed equivariance~\citep{wang2022approximately,pertigkiozoglou2024improving,manolache2025learning,elhag2025relaxed}, while leaving the optimizer itself largely unchanged, even though the optimizer shapes both convergence and which solutions are reached~\citep{pascanu2025optimizersqualitativelyaltersolutions}.

Only recently has the optimizer choice become an explicit part of this discussion. Matrix structured optimizers such as Muon~\citep{jordan2024muon} can outperform Adam~\citep{adam} on equivariant and geometric models~\citep{stupariu2026how}, although comparisons on interatomic potentials find that the gains depend on the architecture and optimizer~\citep{harari2026beyond}. Why these optimizers behave differently on equivariant networks remains mostly unclear.

In this work, we study why matrix structured optimizers such as Muon can outperform Adam on equivariant networks.
Parameter sharing can change SGD's update scales through gradient accumulation, an effect Adam largely compensates for.
A separate issue remains at the matrix level: Adam rescales entries individually without controlling each block's effective step size, so a single learning rate can produce substantially different updates across irrep blocks within a layer.
In a controlled setting, this step size mismatch grows with width in an equivariant model but not in a matched non-equivariant model.

Our contributions are as follows:
\begin{itemize}
	\item We characterize two distinct effects of equivariant layer structure on optimization: the gradient scaling caused by irrep-dependent parameter sharing, which Adam largely compensates for, and a separate step size mismatch across irrep blocks that persists under Adam.
	\item We study spectral and root-mean-square normalization of optimizer updates, including their interaction with the learning rate and Adam's moment coefficients.
	\item We evaluate spectral normalization in a controlled $\mathrm{SO}(3)$-equivariant model with a matched non-equivariant model, and both normalizations in two interatomic potential architectures on rMD17~\citep{md17} and MD22~\citep{md22}, molecular property prediction on QM9~\citep{Rud+2012,Ram+2014}, and charged particle dynamics~\citep{Kip+2018,pmlr-v139-satorras21a}.
\end{itemize}

Block normalization often improves training without relaxing the model's equivariance constraints.
These gains challenge the view that equivariant networks are intrinsically difficult to train and call for treating optimization as part of their design.

\section{Related Work}

\paragraph{Equivariance and training.}
How inputs and intermediate features transform under a symmetry determines which maps an equivariant layer can learn and where its weights are shared.
\citet{petrache2023approximationgeneralization} quantify the approximation-generalization trade-off under approximate or partial equivariance, relating the symmetry of the model to that of the data.
\citet{xie2025a} show that, under particular conditions, the geometry of an equivariant parameter space can prevent learning a global minimum.
\citet{nordenfors2025optimization} compare equivariant architectures with data augmentation. Under suitable conditions, the two approaches share equivariant stationary points, although those points can differ in stability. One response has been to relax exact equivariance while retaining symmetry as an inductive bias.
Soft priors and learned relaxations allow a model to accommodate imperfect symmetries~\citep{NEURIPS2021_fc394e99,wang2022approximately,ouderaa2022relaxing}.
Relaxed equivariant layers can also represent symmetry breaking, including cases where an output should not preserve every symmetry of an individual input~\citep{hofgard2024relaxed,kaba2023symmetry}.
Other approaches relax the constraint during training or encourage equivariance through the objective~\citep{pertigkiozoglou2024improving,manolache2025learning,elhag2025relaxed}.

\paragraph{Optimizers for equivariant models.}
\citet{harari2026beyond} compare Adam~\citep{adam} with Muon~\citep{jordan2024muon} and SOAP~\citep{vyas2025soap} on equivariant interatomic potential models, finding strong results for SOAP and Muon.
\citet{li2026dpa4pushingaccuracycostfrontier} train DPA4 with a hybrid optimizer that applies Muon separately to the channel mixing matrices of each degree and uses Adam updates for other parameters, while~\citet{chen2026protein} apply Muon to two-dimensional weights and Adam to the remaining parameters of equivariant protein fold classification models, reporting substantially improved convergence with Muon. \citet{zhang2026polaradamwdisentanglingspectralcontrol} introduces PolarAdamW, which applies Muon's polar transform to AdamW's preconditioned update, showing how a $\mathrm{SO}(3)$ model's updates respond to changes of basis among irrep channels. \citet{stupariu2026how} find that Adam and Muon reach different performance, loss geometry, and learned weights across equivariant and geometric models, showing how optimizers can shape the solutions reached~\citep{pascanu2025optimizersqualitativelyaltersolutions}. Optimizers such as Muon are becoming a more common choice for training equivariant networks.
Much of the evidence comes from performance comparisons and observed differences in learned weights and loss geometry, which document the gains but leave their causes only partly understood.

\paragraph{Matrix update magnitude.}
Several works control the size of a matrix update during optimization.
\citet{yang2023spectral} show that the spectral norms of weight matrices and their updates must scale appropriately with model width.
\citet{large-scalable} combine norms for components such as linear maps and convolutions into a network-wide norm used to rescale optimizer updates.
Methods such as LARS~\citep{you2017largebatchtrainingconvolutional} and LAMB~\citep{You2020Large} scale each layer's update by the ratio of its weight norm to the norm of the proposed update. \citet{Shazeer2018AdafactorAL} use the root-mean-square size of an update to limit large steps from old second moment estimates. These works study update size more generally, without examining the parameter structure imposed by equivariance. Accounting for this structure raises a more specific question: how should the effective step size of each irrep block be controlled within a layer? We first describe how equivariant layers organize and share parameters, then examine how this structure affects optimizer updates.

\section{Background}

\begin{wrapfigure}{r}{0.35\linewidth}
	\vspace{-0.6\baselineskip}
	\centering

	\begin{subfigure}{\linewidth}
		\centering
		\includegraphics[width=\linewidth]{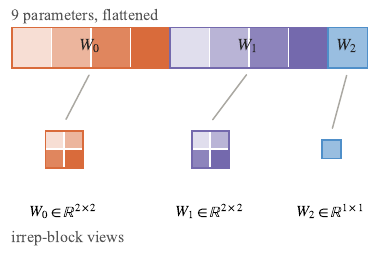}
		\caption{Flat storage and irrep-block views.}
		\label{fig:irrep-blocks-storage}
	\end{subfigure}

	\vspace{0.4em}

	\begin{subfigure}{\linewidth}
		\centering
		\includegraphics[width=\linewidth]{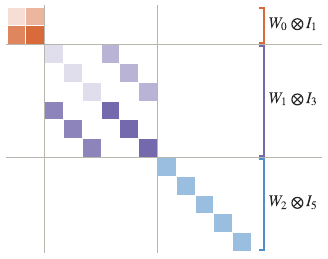}
		\caption{The corresponding expanded map.}
		\label{fig:irrep-blocks-map}
	\end{subfigure}

	\caption{Stored and expanded views of an equivariant linear layer.
		Equal colors indicate tied weights.}
	\label{fig:irrep-blocks}
	\vspace{-2\baselineskip}
\end{wrapfigure}

\paragraph{Equivariance.}
A function $f : X \to Y$ is equivariant to a group $G$ if
$f(\rho_X(g)x) = \rho_Y(g)f(x)$ for all $g \in G$ and $x \in X$, where $\rho_X$ and $\rho_Y$ describe how the group acts on the input and output.
Thus, transforming the input before applying $f$ is equivalent to transforming its output afterward.
Compositions of equivariant functions remain equivariant, allowing an equivariant network to be built one layer at a time.
Convolutional networks are a standard example: sharing a filter across spatial positions makes convolution translation equivariant, so translating the input translates the resulting feature maps.
In this work, we focus on $G=O(3)$, the group of rotations and reflections in three dimensions.
An element of $O(3)$ transforms point coordinates and polar vector features by the same orthogonal matrix.
Even scalar features remain unchanged, whereas pseudoscalars change sign under inversion.
For physical predictions, energy is an invariant scalar, while forces are polar vectors that transform with the input coordinates.

\paragraph{Structure of equivariant layers.}
Many neural networks equivariant to $O(3)$ organize their features into irreducible representations, or irreps, as in e3nn~\citep{e3nn} (e.g., through spherical harmonics~\citep{spherical_harmonics,thomas18}).
An irrep of degree $l$ has $2l+1$ components that transform together, with $l=0$ and $l=1$ corresponding to scalars and vectors.
The multiplicity $m_l$ of an irrep counts its copies, or channels.
For example, $m_1=2$ denotes two vector channels.
For $O(3)$, an irrep also has a parity, which we suppress in the notation.
The index $l$ implicitly includes parity, and equivariant linear layers only mix channels with the same degree and parity.
At each degree $l$, an equivariant linear layer learns channel mixing matrices $W_l \in \mathbb{R}^{m_l^{\mathrm{out}}\times m_l^{\mathrm{in}}}$.
Equivariance requires the same matrix to act on all $2l+1$ components, giving the expanded map $W_l \otimes I_{2l+1}$.
We refer to this channel mixing matrix as an irrep block.
e3nn additionally multiplies each irrep block by a constant determined by the number of input channels.
We measure and normalize updates to $W_l$ without this constant.
Figure~\ref{fig:irrep-blocks} shows a layer in its stored and expanded forms, omitting this additional scaling.
The example has $m_0=m_1=2$ and $m_2=1$ at both input and output, so $W_0$ and $W_1$ contain four parameters each, whereas $W_2$ contains only one.
In the expanded map, each entry is repeated one, three, and five times, respectively.

\section{Block Updates}
\label{subsec:adam_block_update_magnitudes}

\paragraph{Optimizer updates.}
Let $G_l^{(t)}$ denote the gradient of $W_l$ at optimization step $t$.
Stochastic gradient descent applies
$\Delta W_l^{(t)}=-\eta_t G_l^{(t)}$, where $\eta_t$ is the learning rate at step $t$.
Equivariant implementations generally store the irrep blocks $W_l$ and reuse their entries across irrep components, rather than storing the full operators $W_l\otimes I_{2l+1}$.
The matrices may be stored separately or packed into a flat parameter vector, as in e3nn.
Because the same weight is shared for all $2l+1$ components, backpropagation sums their contributions to compute the gradient.
Assuming that these contributions are identical, both the gradient and the SGD update are $2l+1$ times larger than those from one component alone.
Blocks within one layer can therefore receive different effective step sizes simply because their parameters are shared different numbers of times.
Normalizing each block update removes this multiplicative factor, as we describe in Section~\ref{sec:block_normalization}.
Nevertheless, our main focus is Adam, which is commonly used to train equivariant networks.
Unlike SGD, Adam can compensate for the increase in gradient magnitude caused by shared parameters.
To see this, let $g_s$ denote the contribution from one irrep component to the gradient of the shared weight $(W_l)_{ij}$ at step $s\in\{1,\ldots,t\}$.
Assume that at every training step $s$, all $2l+1$ components contribute the same value $g_s$ to the gradient of the shared weight $(W_l)_{ij}$.
The accumulated gradient is then $(2l+1)g_s$.
Writing $\widehat m_t$ and $\widehat v_t$ for the bias-corrected moments of the gradient history, Adam's update becomes
\[
	\begin{aligned}
		\Delta W_{l,ij}^{(t)}
		 & =
		-\eta_t
		\frac{\frac{1-\beta_1}{1-\beta_1^t}
			\sum_{s=1}^{t}\beta_1^{t-s}(2l+1)g_s}
		{\sqrt{\frac{1-\beta_2}{1-\beta_2^t}
				\sum_{s=1}^{t}\beta_2^{t-s}\big((2l+1)g_s\big)^2}
			+\varepsilon}
		\\
		 & =
		-\eta_t
		\frac{(2l+1)\widehat m_t}
		{(2l+1)\sqrt{\widehat v_t}+\varepsilon}
		\approx
		-\eta_t\frac{\widehat m_t}{\sqrt{\widehat v_t}}.
	\end{aligned}
\]
The factor $2l+1$ cancels when the numerical stabilizer $\varepsilon$ is negligible.
Adam largely compensates for this gradient scaling, but still rescales weights individually rather than controlling each block's effective step size. Therefore, a single learning rate can still produce different effective step sizes across blocks.
We examine this remaining mismatch next.

\paragraph{Effective step sizes.}
For an ordinary linear layer, the learning rate scales a single matrix update.
An equivariant layer instead contains one irrep block $W_l$ for each irrep type, all using the same learning rate.
Let $U_l^{(t)}$ be the optimizer's proposed update.
For any matrix norm,
$\lVert\Delta W_l^{(t)}\rVert_\star=\eta_t\lVert U_l^{(t)}\rVert_\star$.
Adam rescales the entries of $U_l^{(t)}$ independently, without using block boundaries or controlling the norm of the complete matrix.
The norms can therefore differ across irrep blocks, so the same learning rate gives the blocks different effective step sizes $\lVert\Delta W_l^{(t)}\rVert_\star$. We call this the step size mismatch.
Choosing $\eta_t$ to keep the largest update stable may make the other updates unnecessarily small, while increasing it for those blocks may make the largest update unstable.
Changing the learning rate cannot resolve this mismatch because it scales every block by the same factor.
We test whether making the learning rate directly set each block's effective step size improves training.

\subsection{Block Normalization}
\label{sec:block_normalization}

\paragraph{Block normalization.}

We normalize Adam's updates separately for each irrep block $W_l$ so that the learning rate directly sets each block's effective step size.
Adam proposes
\[
	U_l^{(t)}
	=
	\frac{\widehat m_l^{(t)}}{\sqrt{\widehat v_l^{(t)}}+\varepsilon},
\]
where $\widehat m_l^{(t)}$ and $\widehat v_l^{(t)}$ contain the bias-corrected first and second moment estimates for the gradients of $W_l$, and all operations are entrywise.
We then set
\[
	\Delta W_l^{(t)}
	=
	\begin{cases}
		-\eta_t \dfrac{U_l^{(t)}}{\lVert U_l^{(t)}\rVert_\star},
		   & U_l^{(t)}\neq 0, \\[6pt]
		0, & U_l^{(t)}=0,
	\end{cases}
\]
where $\lVert\cdot\rVert_\star$ is the norm used for normalization.
Every nonzero update therefore satisfies $\lVert\Delta W_l^{(t)}\rVert_\star=\eta_t$.
This sets each nonzero block update's norm to the learning rate, while leaving Adam's moment-update rule unchanged and preserving the direction of each proposed block update.

\paragraph{Spectral normalization.}
We consider two choices of norm.
The first is the spectral norm, which measures a matrix by its largest singular value:
\[
	\lVert U\rVert_2=\sigma_{\max}(U).
\]
Dividing by this value makes the largest singular value of each block update equal to the learning rate, while preserving its singular value ratios.
For small irrep blocks, we compute $\sigma_{\max}$ by singular value decomposition in float64.
For GotenNet's wider matrices, repeated squaring of the Gram matrix $U^\top U$ gives a high-order Schatten norm, an upper bound on the spectral norm~\citep{horn2012matrix}.
Appendix~\ref{app:spectral-norm-approximation} gives the approximation and its implementation details.

\paragraph{Root-mean-square normalization.} A simpler alternative uses the root-mean-square (RMS) norm,
\[
	\lVert U\rVert_{\mathrm{RMS}}
	=
	\sqrt{\frac{1}{mn}\sum_{i=1}^{m}\sum_{j=1}^{n}U_{ij}^{2}}
	=
	\frac{\lVert U\rVert_{\mathrm F}}{\sqrt{mn}},
\]
for an $m\times n$ block.
This makes the RMS of each block update equal to the learning rate without computing singular values.
Its spectral norm, however, remains dependent on the update:
\[
	\lVert\Delta W\rVert_2
	=
	\eta_t\sqrt{mn}\,
	\frac{\lVert U\rVert_2}{\lVert U\rVert_{\mathrm F}}.
\]
For example, normalizing $U=I_k$ gives a spectral norm of $\eta_t\sqrt{k}$, whereas normalizing a $k \times k$ matrix of ones gives $\eta_t k$.
These examples have the same RMS after normalization but different spectral norms.
Spectral normalization controls the largest singular value, whereas RMS normalization controls the scale of the update entries.
RMS normalization can improve training without equalizing the spectral norms of the block updates.
Zero updates remain zero under either normalization.

\begin{table}[t]
	\centering
	\caption{Test loss on the synthetic task (mean $\pm$ std over eight seeds; lower is better). \textit{spectral} applies spectral block normalization with Adam's default moment coefficients. \textit{Transferred mom.} applies to unnormalized Adam the coefficients selected with normalization, while \textit{both} combines those coefficients with normalization. Bold marks the best result per width and underline the second best. Spectral normalization outperforms Muon from width 128 onward.}
	\label{tab:toy-width}
	\small
	\setlength{\tabcolsep}{6pt}
	\renewcommand{\arraystretch}{1.1}
	\begin{tabular}{rccccc}
		\toprule
		Width & Adam              & + spectral                    & + transferred mom. & + both                        & Muon                          \\
		\midrule
		16    & $0.675 \pm 0.021$ & $\underline{0.654 \pm 0.031}$ & $0.705 \pm 0.028$  & $0.674 \pm 0.017$             & $\mathbf{0.634 \pm 0.022}$    \\
		32    & $0.557 \pm 0.022$ & $0.564 \pm 0.031$             & $0.554 \pm 0.019$  & $\underline{0.537 \pm 0.022}$ & $\mathbf{0.513 \pm 0.017}$    \\
		64    & $0.449 \pm 0.014$ & $0.419 \pm 0.019$             & $0.479 \pm 0.031$  & $\mathbf{0.376 \pm 0.017}$    & $\underline{0.394 \pm 0.022}$ \\
		128   & $0.330 \pm 0.017$ & $\underline{0.279 \pm 0.014}$ & $0.399 \pm 0.021$  & $\mathbf{0.225 \pm 0.006}$    & $0.300 \pm 0.018$             \\
		192   & $0.265 \pm 0.019$ & $\underline{0.220 \pm 0.017}$ & $0.257 \pm 0.015$  & $\mathbf{0.156 \pm 0.008}$    & $0.249 \pm 0.009$             \\
		256   & $0.230 \pm 0.008$ & $\underline{0.178 \pm 0.014}$ & $0.220 \pm 0.008$  & $\mathbf{0.121 \pm 0.004}$    & $0.224 \pm 0.010$             \\
		\bottomrule
	\end{tabular}
\end{table}

\section{Empirical Evaluation}
\label{sec:empirical}

We study how equivariant structure affects optimizer updates and whether controlling their effective step sizes improves training.
Our experiments address four questions:
\emph{\textbf{RQ1:}} How do effective step sizes differ across irrep blocks, and how do these differences change with model width?
\emph{\textbf{RQ2:}} Does block normalization improve optimization and predictive performance across equivariant tasks?
\emph{\textbf{RQ3:}} How does normalization interact with the learning rate and Adam's moment coefficients?
\emph{\textbf{RQ4:}} Which parts of an equivariant architecture benefit from different optimizer choices?

\paragraph{Models and datasets.} We first evaluate a toy controlled $\mathrm{SO}(3)$-equivariant model against a matched non-equivariant network, followed by three realistic applications.
For molecular force prediction, we use a NequIP model~\citep{batzner2022nequip} implemented in e3nn~\citep{e3nn}, and GotenNet~\citep{aykent2025gotennet} on the rMD17~\citep{md17} and MD22~\citep{md22} molecular datasets.
We further evaluate GotenNet on the QM9 molecular properties dataset~\citep{Rud+2012,Ram+2014}, and a NequIP model on charged particle dynamics simulations~\citep{Kip+2018,pmlr-v139-satorras21a}.

\paragraph{Protocol.}
We compare Adam, spectral and root-mean-square block normalization, and Muon using the same data splits and number of epochs.
We search learning rates and momentum coefficients for each method, selecting hyperparameters and checkpoints using validation performance.
Runs use cosine learning rate decay and no weight decay.
We report means and standard deviations over multiple seeds.
Architectures, hyperparameter grids, seed counts, and implementation details are provided in Appendix sections~\ref{pytorch-implementation} and~\ref{app:experimental-details}.

\subsection{Synthetic dataset}

\begin{wrapfigure}{r}{0.42\linewidth}
	\vspace{-1\baselineskip}
	\centering
	\includegraphics[width=\linewidth]{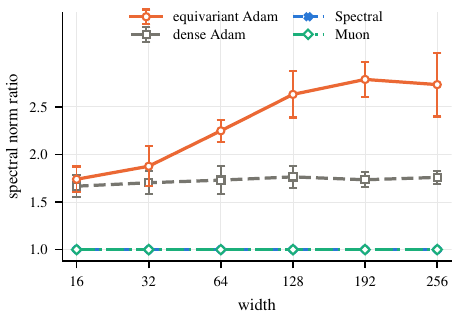}
	\caption{Ratio between the largest and smallest spectral norms of the matrix updates at Adam's first update. The mismatch increases with width in the equivariant model but remains nearly constant in the matched dense MLP. Spectral and Muon have a ratio of one by construction.
		Mean and standard deviations over eight seeds.
	}
	\label{fig:toy-step-magnitudes}
	\vspace{-4\baselineskip}
\end{wrapfigure}

\paragraph{Synthetic dataset.} To answer \emph{\textbf{RQ1}},
we consider $\mathrm{SO}(3)$-invariant regression on 8192 training point clouds, each containing 16 random 3D points, with 512 validation and 2048 test point clouds.
The task is to predict the largest eigenvalue of each cloud's second-moment matrix, a scalar unchanged by rotation.
We compare a $\mathrm{SO}(3)$-equivariant network against a dense MLP with the same hidden dimension.
We train widths $16$, $32$, $64$, $128$, $192$, and $256$ for 1000 steps, searching the learning rate, initialization scale, and moment coefficients over eight seeds.
We compare Adam, Adam with spectral block normalization, Adam with tuned moment coefficients, their combination, and Muon.

Figure~\ref{fig:toy-step-magnitudes} measures the variation in the spectral norms of the block updates at Adam's first step. We compute
\[
	R =
	\frac{\max_{b\in\mathcal B_t}\lVert\Delta W_b^{(t)}\rVert_2}
	{\min_{b\in\mathcal B_t}\lVert\Delta W_b^{(t)}\rVert_2},
\]

where $\mathcal B_t$ contains the nonzero matrix updates in all three hidden layers; input and readout matrices are excluded. We take the ratio over all hidden layers because the non-equivariant model has only one matrix per layer; Figure~\ref{fig:toy-within-layer} in the appendix shows the ratio within each layer. Width $m$ gives the equivariant model $m$ scalar and $m$ vector channels, matching the non-equivariant model at hidden dimension $4m$. In these architectures, $R$ increases with width for the equivariant model but remains nearly constant for the non-equivariant model. This may make the learning rate harder to choose, since stabilizing the largest updates can slow the others. Exact spectral normalization and Muon's exact polar update give $R=1$ by construction.

To examine whether removing the mismatch improves training, Table~\ref{tab:toy-width} compares the test losses of the best checkpoints.
In the narrower models, spectral normalization performs similarly to Adam.
As width increases, normalization provides larger improvements and eventually outperforms Muon.
Combining normalization with tuned moment coefficients further improves the wider models, whereas moment tuning alone provides no consistent benefit across widths.
Having found that normalization helps in this synthetic setting, we next evaluate it on more realistic models and datasets.

\subsection{Interatomic Potentials}

\begin{table}[t]
	\centering
	\caption{Test force MAE (kcal/mol/\AA, $\downarrow$) for NequIP on rMD17 and MD22. Results show the mean and std over three seeds. For each optimizer, we select learning rates, moment coefficients, and checkpoints using validation error. Bold: best result; underline: second best.}
	\label{tab:force-mae-nequip}
	\small
	\setlength{\tabcolsep}{5pt}
	\begin{tabular}{clcccc}
		\toprule
		 & Dataset        & Adam              & Spectral          & RMS                           & Muon                          \\
		\midrule
		\multirow{10}{*}{\rotatebox[origin=c]{90}{\scriptsize rMD17}}
		 & ethanol        & $0.366 \pm 0.006$ & $0.375 \pm 0.002$ & $\underline{0.319 \pm 0.009}$ & $\mathbf{0.262 \pm 0.001}$    \\
		 & malonaldehyde  & $0.542 \pm 0.029$ & $0.545 \pm 0.010$ & $\underline{0.474 \pm 0.015}$ & $\mathbf{0.408 \pm 0.006}$    \\
		 & benzene        & $0.159 \pm 0.006$ & $0.135 \pm 0.004$ & $\underline{0.118 \pm 0.002}$ & $\mathbf{0.094 \pm 0.003}$    \\
		 & uracil         & $0.487 \pm 0.019$ & $0.473 \pm 0.011$ & $\underline{0.382 \pm 0.015}$ & $\mathbf{0.357 \pm 0.002}$    \\
		 & toluene        & $0.489 \pm 0.037$ & $0.451 \pm 0.007$ & $\underline{0.392 \pm 0.009}$ & $\mathbf{0.360 \pm 0.004}$    \\
		 & salicylic acid & $0.642 \pm 0.006$ & $0.637 \pm 0.008$ & $\underline{0.564 \pm 0.007}$ & $\mathbf{0.478 \pm 0.008}$    \\
		 & naphthalene    & $0.477 \pm 0.023$ & $0.421 \pm 0.007$ & $\underline{0.356 \pm 0.009}$ & $\mathbf{0.321 \pm 0.002}$    \\
		 & paracetamol    & $0.681 \pm 0.012$ & $0.690 \pm 0.008$ & $\underline{0.597 \pm 0.013}$ & $\mathbf{0.542 \pm 0.009}$    \\
		 & aspirin        & $0.817 \pm 0.034$ & $0.824 \pm 0.002$ & $\underline{0.735 \pm 0.005}$ & $\mathbf{0.679 \pm 0.005}$    \\
		 & azobenzene     & $0.633 \pm 0.028$ & $0.622 \pm 0.009$ & $\underline{0.524 \pm 0.012}$ & $\mathbf{0.466 \pm 0.007}$    \\
		\cmidrule{2-6}
		\multirow{2}{*}{\rotatebox[origin=c]{90}{\scriptsize MD22}}
		 & Ac-Ala3-NHMe   & $1.008 \pm 0.090$ & $0.962 \pm 0.005$ & $\mathbf{0.867 \pm 0.019}$    & $\underline{0.878 \pm 0.003}$ \\
		 & stachyose      & $1.782 \pm 0.055$ & $1.478 \pm 0.009$ & $\mathbf{1.386 \pm 0.016}$    & $\underline{1.419 \pm 0.020}$ \\
		\bottomrule
	\end{tabular}
\end{table}

\begin{figure}[t]
	\centering
	\includegraphics[width=\textwidth]{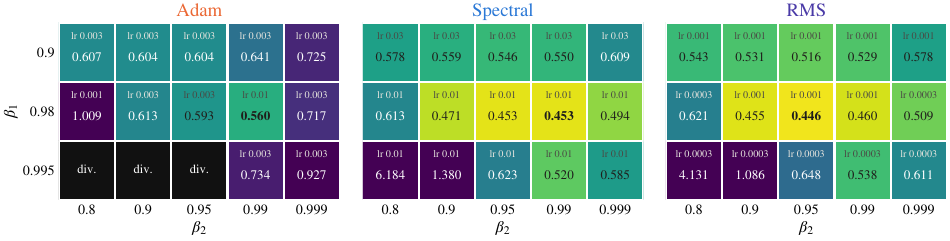}
	\caption{Test force MAE (kcal/mol/\AA, $\downarrow$) for GotenNet on rMD17 aspirin across moment coefficients $(\beta_1,\beta_2)$. Results are averaged over three seeds, with 3000 steps per run. Learning rates are selected using validation error and shown above each result. Bold marks the best configuration per method; black indicates divergence at all tested learning rates.}
	\label{fig:moment-grid}
\end{figure}
\begin{figure}[t]
	\centering
	\begin{subfigure}[t]{0.36\textwidth}
		\centering
		\includegraphics[width=\linewidth]{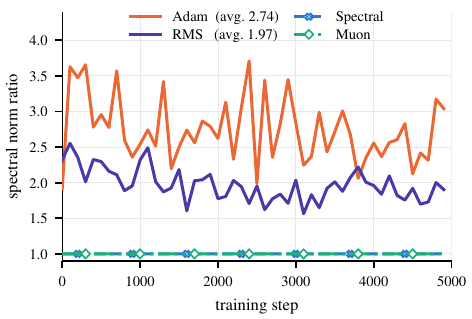}
		\caption{Ratio of spectral norms.}
		\label{fig:block-step-normalization}
	\end{subfigure}%
	\hfill%
	\begin{subfigure}[t]{0.62\textwidth}
		\centering
		\includegraphics[width=\linewidth]{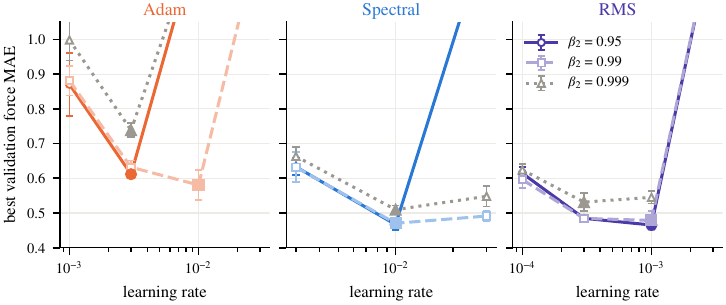}
		\caption{Validation force MAE ($\downarrow$) against the learning rate at $\beta_1 = 0.98$.}
		\label{fig:lr-curves}
	\end{subfigure}
	\caption{Results on rMD17 aspirin. (a) Ratio between the largest and smallest spectral norms of irrep block updates; legend averages are over the displayed training steps. (b) GotenNet validation force MAE versus learning rate, with $\beta_1=0.98$ and varying $\beta_2$, averaged over three seeds. Curves leaving the plot indicate divergence at the next tested rate.}
	\label{fig:mismatch-and-lr}
\end{figure}

\paragraph{Effective step sizes during training.}
We now turn to molecular data to test whether the mismatch observed in the synthetic setting persists during training in interatomic potential models (\emph{\textbf{RQ1}}).
Computing the spectral norm by SVD for every block at every step becomes costly as the number and size of the layers increase.
For GotenNet, we use the spectral norm approximation described in Section~\ref{sec:block_normalization}.
We also evaluate RMS normalization as a simpler alternative that does not require computing singular values.
Figure~\ref{fig:block-step-normalization} plots the ratio between the largest and smallest spectral norms of the block updates during NequIP training on rMD17 aspirin.
For Adam, this ratio remains high throughout training.
Spectral normalization keeps it at one, while RMS normalization reduces the differences between blocks without fully equalizing their magnitudes.

\paragraph{Molecular force prediction.}
To answer \emph{\textbf{RQ2}}, we examine whether normalization improves prediction accuracy.
For NequIP, Table~\ref{tab:force-mae-nequip} shows that RMS normalization achieves lower mean test force error than both Adam and spectral normalization on every dataset.
It also gives the lowest errors on MD22, while Muon performs best on rMD17.
The GotenNet results in Appendix Table~\ref{tab:force-mae-gotennet} show a different pattern.
Both normalizations consistently improve Adam, but Muon achieves the lowest error on every dataset, and spectral normalization is generally the stronger of the two.
The preferred normalization therefore depends on the interatomic potential model, and we investigate potential reasons for Muon's remaining advantage later in this section.

\paragraph{Learning rate and moment coefficients.}
To answer \emph{\textbf{RQ3}}, we vary the learning rate and Adam's moment coefficients for GotenNet on rMD17 aspirin, training each model for 3000 steps.
Figure~\ref{fig:moment-grid} compares the resulting test errors, with the learning rate selected separately for each pair of moment coefficients.
Both normalizations improve Adam at every tested pair.
The models also remain trainable at several moment settings where Adam diverges, although some of these settings still produce high errors.
The learning rate search in Figure~\ref{fig:lr-curves} shows a similar difficulty for Adam: for some moment settings, fewer of the tested rates lead to stable training.
Normalization therefore broadens the range of useful hyperparameter settings and makes the search less prone to divergent runs, making it easier to find well-performing models with a limited tuning budget.

\paragraph{Other base optimizers.}
\begin{wraptable}{r}{0.48\textwidth}
	\centering
	\vspace{-1.2em}
	\caption{Test force MAE (kcal/mol/\AA, $\downarrow$) on rMD17 aspirin with NequIP for Adam and SGD, with and without normalization. Subscripts give the standard deviation over three seeds.}
	\label{tab:sgd}
	\footnotesize
	\setlength{\tabcolsep}{4pt}
	\begin{tabular}{lccc}
		\toprule
		Base & None                & Spectral            & RMS                 \\
		\midrule
		Adam & $0.817_{\pm 0.034}$ & $0.824_{\pm 0.002}$ & $0.735_{\pm 0.005}$ \\
		SGD  & $1.534_{\pm 0.019}$ & $0.800_{\pm 0.008}$ & $0.761_{\pm 0.008}$ \\
		\bottomrule
	\end{tabular}
\end{wraptable}
Block normalization rescales the update proposed by the base optimizer, so it applies to SGD as well as to Adam.
SGD with momentum accumulates gradients without Adam's entrywise rescaling, leaving its updates sensitive to the gradient scaling discussed in \Cref{subsec:adam_block_update_magnitudes}.
\Cref{tab:sgd} compares the two base optimizers on rMD17 aspirin with NequIP.
Without normalization, SGD has substantially higher test error than Adam. Both spectral and RMS normalization nearly halve its error.
Block normalization removes the gradient scaling from parameter sharing, which Adam removes implicitly through its second moment estimate, and with it the two base optimizers reach similar test errors.

\begin{figure}[t]
	\centering
	\includegraphics[width=\textwidth]{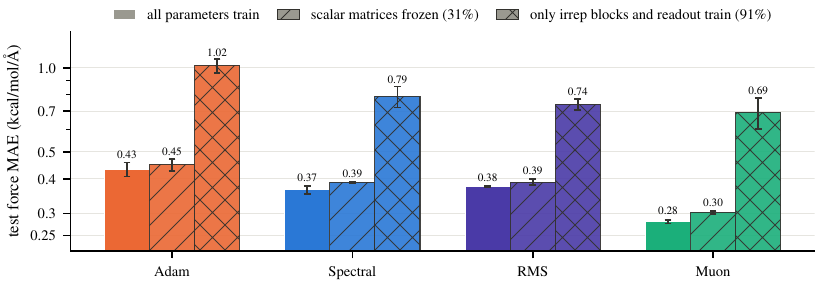}
	\caption{Test force MAE (kcal/mol/\AA, $\downarrow$) for GotenNet on rMD17 aspirin with all parameters trained, scalar matrices frozen, or only the irrep blocks and readout trained. The number in parentheses is the percentage of parameters frozen at their initial values. Mean and std over three seeds.}
	\label{fig:freeze}
\end{figure}

\paragraph{Ablation study.}
Muon's larger advantage on GotenNet raises the possibility that it benefits ordinary scalar matrices more than irrep blocks (\emph{\textbf{RQ4}}).
We investigate this on rMD17 aspirin by freezing subsets of parameters at initialization.
Figure~\ref{fig:freeze} shows that freezing scalar matrices, including attention projections and the readout, leaves the optimizer differences largely unchanged.
We next train only the irrep blocks and readout, freezing everything else.
This includes the scalar networks that GotenNet uses to scale its steerable features, along with the atom embeddings and feature normalization parameters.
Although all methods become less accurate, both normalizations retain their improvement over Adam.
RMS normalization reaches a similar mean error to Muon, whose results vary more across seeds.
Muon's larger advantage in the full model may depend on how the irrep blocks are optimized together with other components.
We leave a closer study of these interactions to future work.

\subsection{Molecular Properties and Particle Dynamics}
Finally, we test whether the benefits of block normalization extend beyond molecular force prediction (\emph{\textbf{RQ2}}).
We consider invariant molecular property prediction on QM9~\citep{Rud+2012,Ram+2014} and equivariant prediction of future positions in charged particle dynamics~\citep{Kip+2018,pmlr-v139-satorras21a}.

\begin{figure}[t]
	\centering
	\begin{minipage}[c]{0.52\textwidth}
		\centering
		\captionof{table}{Test MAE ($\downarrow$) on QM9 with GotenNet, scaled by each target's training set standard deviation and multiplied by 100 (mean $\pm$ std over three seeds). Block normalization gives the lowest error on every target. Bold: best per row; underline: second best.}
		\label{tab:qm9}
		\scriptsize
		\setlength{\tabcolsep}{3pt}
		\begin{tabular}{lccc}
			\toprule
			                              & Adam                          & Spectral                      & RMS                           \\
			\midrule
			$\mu$                         & $2.095 \pm 0.221$             & $\underline{1.970 \pm 0.050}$ & $\mathbf{1.884 \pm 0.022}$    \\
			$\alpha$                      & $1.710 \pm 0.040$             & $\underline{1.649 \pm 0.021}$ & $\mathbf{1.632 \pm 0.048}$    \\
			$\varepsilon_{\mathrm{HOMO}}$ & $5.614 \pm 0.136$             & $\underline{5.394 \pm 0.073}$ & $\mathbf{5.347 \pm 0.041}$    \\
			$\varepsilon_{\mathrm{LUMO}}$ & $2.975 \pm 0.058$             & $\mathbf{2.816 \pm 0.038}$    & $\underline{2.827 \pm 0.036}$ \\
			$\Delta\varepsilon$           & $3.855 \pm 0.059$             & $\mathbf{3.666 \pm 0.046}$    & $\underline{3.678 \pm 0.052}$ \\
			$\langle R^2 \rangle$         & $\underline{0.522 \pm 0.006}$ & $\mathbf{0.462 \pm 0.014}$    & $0.563 \pm 0.009$             \\
			ZPVE                          & $\underline{0.377 \pm 0.014}$ & $0.381 \pm 0.002$             & $\mathbf{0.363 \pm 0.005}$    \\
			$U_0$                         & $0.469 \pm 0.009$             & $\underline{0.462 \pm 0.003}$ & $\mathbf{0.441 \pm 0.013}$    \\
			$U$                           & $0.464 \pm 0.002$             & $\underline{0.457 \pm 0.007}$ & $\mathbf{0.441 \pm 0.013}$    \\
			$H$                           & $\underline{0.459 \pm 0.006}$ & $0.460 \pm 0.003$             & $\mathbf{0.445 \pm 0.016}$    \\
			$G$                           & $0.500 \pm 0.007$             & $\underline{0.496 \pm 0.004}$ & $\mathbf{0.476 \pm 0.016}$    \\
			$c_v$                         & $1.411 \pm 0.033$             & $\underline{1.379 \pm 0.032}$ & $\mathbf{1.337 \pm 0.016}$    \\
			\midrule
			Average                       & $1.704 \pm 0.044$             & $\underline{1.633 \pm 0.021}$ & $\mathbf{1.619 \pm 0.023}$    \\
			\bottomrule
		\end{tabular}
	\end{minipage}%
	\hfill%
	\begin{minipage}[c]{0.45\textwidth}
		\centering
		\vspace{-2em}
		\includegraphics[width=\linewidth]{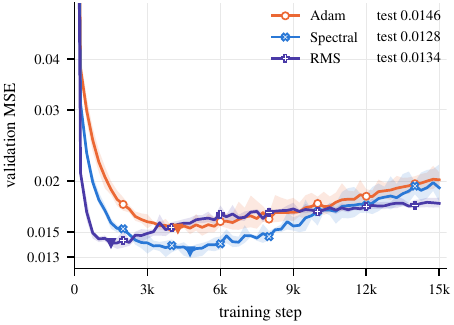}
		\captionof{figure}{Mean and standard deviation of validation MSE ($\downarrow$) for NequIP on charged particle dynamics. Triangles mark the minima of the mean curves. The legend reports mean test MSE at checkpoints selected per run using validation error. All models overfit before the first 5000 steps, with RMS converging faster but Spectral offering the best checkpoints.}

		\label{fig:nbody}
	\end{minipage}
\end{figure}

\paragraph{QM9.}
We train GotenNet for 200k steps to predict all twelve molecular properties jointly, using a shared representation and a separate prediction head for each target.
Table~\ref{tab:qm9} reports normalized test MAE ($\times 100$), allowing errors to be averaged across properties with different units.
Both normalizations improve on Adam on average.
RMS achieves the lowest mean error on most targets, while spectral normalization performs better on a smaller set.
The clearest difference is for $\langle R^2\rangle$, where spectral normalization reduces the error compared with Adam, whereas RMS increases it.
Despite these differences across targets, the two methods reach similar average errors.

\paragraph{Particle dynamics.}
We use a NequIP model with velocity inputs to predict the future positions of five interacting charged particles, training each model for 30k steps.
Both normalizations reduce test MSE compared with Adam.
Figure~\ref{fig:nbody} shows how the training dynamics differ: RMS reduces validation error more quickly at the start, while spectral normalization reaches lower validation error and test MSE after more updates.
In this setting, faster initial progress does not translate into the best predictive performance.
For all three methods, validation error eventually increases with continued training, so we only plot the first 15k steps.

\section{Conclusion, Limitations and Future Work}

We identified and analyzed a step size mismatch in equivariant layers.
Parameter sharing can change SGD's step sizes through gradient accumulation.
Adam reduces sensitivity to gradient scale, but its entrywise adaptation does not control the effective step size of each irrep block.
Spectral and RMS normalization address this mismatch without introducing an additional hyperparameter.
In the toy model, spectral normalization became more beneficial as width increased.
Across molecular force prediction, QM9, and charged particle dynamics, block normalization generally improved Adam, although the preferred normalization depended on the model and task.

These findings connect the structure of equivariant layers to their optimization and show that this connection has practical consequences.
Normalizing block updates improved training without changing the architecture or relaxing its equivariance constraints.
This gives a reason to examine how optimizers handle equivariant structure before attributing training difficulties to equivariance itself.
Muon remains the stronger optimizer on most datasets, so beyond the step size mismatch analyzed here, some other property of Muon must help equivariant networks.

Our evaluation focused on $O(3)$-equivariant models and two choices of normalization.
Future work should examine whether different irrep blocks benefit from different target step sizes and how normalization can be combined with other optimizer mechanisms.
The polar transformation studied in PolarAdamW~\citep{zhang2026polaradamwdisentanglingspectralcontrol} provides one direction, while our freezing ablations point to the importance of studying irrep blocks together with attention, gating, and other components.
Understanding these interactions could guide optimizers designed for equivariant architectures and extend this work to other symmetry groups and other forms of parameter sharing.

\section*{LLM Usage Disclosure}
Large language models assisted with writing and debugging code, discussing experimental designs and results, searching related literature, drafting, reviewing, and polishing the manuscript, and checking implementations and mathematical derivations for possible errors.
The authors reviewed LLM-generated code and ran numerical tests of the core implementations, including checks of update norms, preservation of update directions, optimizer moment calculations, and model equivariance.
The mathematical identities were also checked through derivations and numerical tests.
The research questions, central ideas, overall direction, and experimental protocol were determined by the authors, who made all final experimental decisions and decided which suggestions and interpretations to adopt.
All AI-assisted work was reviewed by the authors, who take responsibility for the paper's content.

\section*{Reproducibility Statement}
Section~\ref{sec:block_normalization} defines block normalization, Algorithm~\ref{alg:block-normalization} in Appendix~\ref{pytorch-implementation} gives its implementation for e3nn, and Appendix~\ref{app:spectral-norm-approximation} describes the spectral norm approximation used for GotenNet.
Appendix~\ref{app:experimental-details} describes the model architectures, data splits and training budgets, and Tables~\ref{tab:experimental-details} and~\ref{tab:selected-hyperparameters} list the searched and selected hyperparameters of every experiment.
Section~\ref{sec:empirical} and Appendix~\ref{app:architectures} fully specify the synthetic task.
All other datasets are public: rMD17, MD22 and QM9 are used with the splits given in Appendix~\ref{app:experimental-details}, and the charged particle dynamics data come from the public generator of \citet{pmlr-v139-satorras21a}.
We will release code, including the optimizer implementations, training scripts and the configuration files of all experiments, together with the final version of the paper.

\bibliographystyle{plainnat}
\bibliography{bibliography}

\clearpage
\appendix

\section{Additional Mathematical Details}
\label{app:mathematical-details}

\paragraph{Irrep features.}
An irreducible representation of $O(3)$ has degree $l$, parity, and $d_l=2l+1$ components.
We suppress parity in the notation, but treat different parities as separate irrep types.
Scalars have $l=0$, vectors have $l=1$, and higher degrees describe features with more components.
With $m_l$ channels of each type, the feature space is
\[
	\mathcal V
	=
	\bigoplus_l
	\left(\mathbb R^{m_l}\otimes\mathbb R^{d_l}\right).
\]
The first factor indexes channels; the second contains the components transformed by rotations and reflections.

\paragraph{Equivariant layers.}
An equivariant linear layer mixes channels of the same irrep using the same coefficients for every component.
To write this operation, let $X_l$ contain one channel per row and one component per column.
The layer computes
\[
	Y_l=W_lX_l,
\]
where $X_l\in\mathbb R^{m_l^{\mathrm{in}}\times d_l}$ and
$W_l\in\mathbb R^{m_l^{\mathrm{out}}\times m_l^{\mathrm{in}}}$.
Each coefficient in $W_l$ therefore multiplies an entire channel, not an individual component.

If we arrange the channels consecutively in a vector, the same operation is represented by $W_l\otimes I_{d_l}$.
The Kronecker product makes the sharing explicit by replacing each coefficient with a scaled identity matrix:
\[
	\begin{bmatrix}
		a & b \\
		c & d
	\end{bmatrix}
	\otimes I_{d_l}
	=
	\begin{bmatrix}
		aI_{d_l} & bI_{d_l} \\
		cI_{d_l} & dI_{d_l}
	\end{bmatrix}.
\]
For example, $bI_{d_l}$ adds the second input channel to the first output channel with coefficient $b$ for every component.
Schur's lemma~\citep{Fulton1991RepresentationTA} requires this structure and prevents a linear layer from mixing different irrep types.
The full weight matrix is therefore block diagonal:
\[
	A=\bigoplus_l\left(W_l\otimes I_{d_l}\right),
\]
with one block per irrep type.

\paragraph{Vector example.}
Consider two vector channels and an irrep block $W_1$,
\[
	v_1=\begin{bmatrix}1\\2\\3\end{bmatrix},
	\qquad
	v_2=\begin{bmatrix}4\\5\\6\end{bmatrix},
	\qquad
	W_1=\begin{bmatrix}2&-1\\0.5&3\end{bmatrix}.
\]
The output vectors are $u_1=2v_1-v_2$ and $u_2=0.5v_1+3v_2$, or equivalently
\[
	\begin{bmatrix}u_1\\u_2\end{bmatrix}
	=
	\underbrace{
		\begin{bmatrix}
			2I_3   & -I_3 \\
			0.5I_3 & 3I_3
		\end{bmatrix}
	}_{W_1\otimes I_3}
	\begin{bmatrix}v_1\\v_2\end{bmatrix}.
\]
The same four coefficients mix the $x$, $y$, and $z$ components.
Rotating both inputs by $R$ gives
$2Rv_1-Rv_2=R(2v_1-v_2)$, and likewise for $u_2$.
An arbitrary learned transformation of the three components would generally break this property.

\paragraph{Gradients of shared weights.}
The same matrix $W_l$ acts on all $d_l$ components, so backpropagation adds their contributions when computing its gradient.
Let $D_l=\partial\mathcal L/\partial Y_l$ be the gradient at the layer's output.
For component $c$, write $x_{l,c}$ for the corresponding column of input features and $\delta_{l,c}$ for the column of output gradients.
Then
\[
	G_l
	=
	\frac{\partial\mathcal L}{\partial W_l}
	=
	D_lX_l^\top
	=
	\sum_{c=1}^{d_l}\delta_{l,c}x_{l,c}^\top.
\]
Each component contributes one outer product, whose rank is at most one.
For one application of the layer, this gives
\[
	\operatorname{rank}(G_l)
	\leq
	\min\left(m_l^{\mathrm{in}},m_l^{\mathrm{out}},d_l\right).
\]
A batch or repeated use of the layer sums several such gradients, so the accumulated gradient can have higher rank.

As discussed in Section~\ref{subsec:adam_block_update_magnitudes}, if every component contributes the same gradient, the sum is $d_l$ times the contribution from one component.
The SGD update grows by the same factor.
Adam responds differently: if this scaling holds at every step, its first moment grows by $d_l$ and its second moment by $d_l^2$.
Taking the square root of the second moment leaves a factor $d_l$ in both numerator and denominator, which cancels to $m/\sqrt{v}$.
This argument also holds with bias correction and assumes a negligible numerical stabilizer.

\paragraph{Effective step sizes.}
SGD and Adam multiply each proposed block update $U_l$ by the learning rate $\eta_t$.
The resulting update $\Delta W_l=-\eta_t U_l$ therefore has norm
$\lVert\Delta W_l\rVert_\star=\eta_t\lVert U_l\rVert_\star$.
Blocks with larger proposed updates take larger steps, even though they use the same learning rate.
Block normalization divides each proposal by its own norm before multiplying by $\eta_t$.
Under this exact normalization rule, each nonzero block update has norm $\eta_t$.

We consider two choices for an $m\times n$ block:
\[
	\lVert U_l\rVert_2=\sigma_{\max}(U_l),
	\qquad
	\lVert U_l\rVert_{\mathrm{RMS}}
	=
	\sqrt{\frac{1}{mn}\sum_{i,j}(U_l)_{ij}^2}
	=
	\frac{\lVert U_l\rVert_{\mathrm F}}{\sqrt{mn}}.
\]
Here, $\lVert\cdot\rVert_{\mathrm F}$ is the Frobenius norm.
Spectral normalization sets the largest singular value of the update to $\eta_t$.
RMS normalization instead sets the root-mean-square size of its entries to $\eta_t$, without computing singular values.

The spectral norm is useful because it bounds how much an update can change the block's output.
For fixed input features $X_l$,
\[
	\lVert\Delta Y_l\rVert_{\mathrm F}
	=
	\lVert\Delta W_lX_l\rVert_{\mathrm F}
	\leq
	\lVert\Delta W_l\rVert_2\lVert X_l\rVert_{\mathrm F}.
\]
This connects the update norm to its effect on the layer.
It does not imply that equal update norms produce equal changes in the loss, since the inputs and surrounding network also matter.
In our experiments, we test whether using a common update norm across blocks improves training.

\paragraph{Block normalization.}
We rescale each nonzero proposed block update $U_l^{(t)}$ as
\[
	\Delta W_l^{(t)}
	=
	-\eta_t
	\frac{U_l^{(t)}}{\lVert U_l^{(t)}\rVert_\star},
	\qquad
	\star\in\{2,\mathrm{RMS}\}.
\]
Zero updates remain zero.
The resulting update has norm $\eta_t$, while retaining its direction and singular value ratios.
The optimizer's moment update rule is unchanged.

\paragraph{Spectral norm approximation.}
\label{app:spectral-norm-approximation}
For bigger blocks, we avoid SVD by estimating the largest singular value.
First, we form $B=U^\top U$, or $B=UU^\top$ if that gives a smaller matrix.
Its eigenvalues are the squared singular values of $U$.
Squaring $B$ twice gives the estimate
\[
	\widehat{\sigma}(U)
	=
	\lVert B^4\rVert_{\mathrm F}^{1/8}
	=
	\left(\sum_i\sigma_i(U)^{16}\right)^{1/16}.
\]
The high power makes the largest singular values dominate the sum.
This estimate is the Schatten-16 norm.
For a nonzero matrix of rank $r$, it satisfies
\[
	\lVert U\rVert_2
	\leq\widehat{\sigma}(U)
	\leq r^{1/16}\lVert U\rVert_2.
\]
We divide the proposed update by this estimate:
\[
	\Delta W=-\eta_t\frac{U}{\widehat{\sigma}(U)}.
\]
Because the estimate is an upper bound, the resulting spectral norm is at most $\eta_t$, rather than exactly $\eta_t$.

\paragraph{Relation to Muon.}
Muon combines gradients using momentum and then changes the singular values of the resulting matrix.
Writing its reduced SVD as $M_l=P_l\Sigma_lQ_l^\top$, the exact polar update is
\[
	\Delta W_l^{\mathrm{Muon}}
	=
	-\eta_tP_lQ_l^\top.
\]
This replaces the singular values by $\eta_t$.
Spectral and RMS normalization instead multiply the proposed update by one scalar, preserving the ratios between its singular values.
Exact spectral normalization and the exact polar update therefore have the same largest singular value, but generally produce different update directions.
RMS normalization fixes the size of the entries and doesn't necessarily share the spectral norm.
Except on the synthetic task, Muon approximates the polar transformation with five Newton-Schulz iterations, so its singular values are not exactly equal.

\section{PyTorch Implementation}
\label{pytorch-implementation}

Algorithm~\ref{alg:block-normalization} shows the block normalization pseudocode for e3nn linear weights.
Adam computes its update and moment estimates as usual.
We then split the flat update into irrep blocks, reshape each into a matrix, and rescale each separately. The code for the paper will be made publicly available together with the final version of the paper.
\renewcommand{\lstlistingname}{Algorithm}

\begin{lstlisting}[
    language=Python,
    basicstyle=\ttfamily\footnotesize,
    frame=single,
    breaklines=true,
    caption={Block normalization for e3nn linear weights.},
    label={alg:block-normalization}
]
def irrep_blocks(model):
    for layer in model.modules():
        if not isinstance(layer, o3.Linear):
            continue

        offset, blocks = 0, []
        for instruction in layer.instructions:
            shape = tuple(instruction.path_shape)
            end = offset + shape[0] * shape[1]
            if end > offset:
                blocks.append((slice(offset, end), shape))
            offset = end

        yield layer.weight, blocks


@torch.no_grad()
def block_normalized_step(model, adam):
    specs = list(irrep_blocks(model))
    old = {id(weight): weight.clone()
           for weight, _ in specs}

    adam.step()

    for weight, blocks in specs:
        eta = current_learning_rate(adam, weight)
        update = weight - old[id(weight)]

        for indices, shape in blocks:
            matrix = update[indices].reshape(shape)
            norm = torch.linalg.matrix_norm(matrix, 2)

            if norm > 0:
                update[indices] = (
                    eta * matrix / norm
                ).reshape(-1)

        weight.copy_(old[id(weight)] + update)
\end{lstlisting}

\clearpage
\paragraph{Computational cost.}
Table~\ref{tab:step-time} reports the time of one optimizer step on rMD17 aspirin.
The exact SVD costs 31\,ms per step on NequIP's small irrep blocks but 470\,ms on GotenNet's wider matrices, so we use repeated squaring for GotenNet.
Repeated squaring costs about as much as Muon's Newton-Schulz iteration, and RMS normalization is the cheapest variant.

\begin{table}[t]
	\centering
	\caption{Time of one optimizer step on rMD17 aspirin (batch size 32, float32, one RTX 5090), in milliseconds.}
	\label{tab:step-time}
	\small
	\setlength{\tabcolsep}{6pt}
	\begin{tabular}{lcc}
		\toprule
		Update rule                 & NequIP & GotenNet \\
		\midrule
		Adam                        & 0.2    & 0.6      \\
		RMS                         & 0.9    & 3.3      \\
		Spectral, repeated squaring & 2.9    & 12.8     \\
		Muon                        & 2.3    & 11.8     \\
		Spectral, exact SVD         & 31.4   & 470.2    \\
		\bottomrule
	\end{tabular}
\end{table}

\section{Experimental Details}
\label{app:experimental-details}

\subsection{Model Architectures}
\label{app:architectures}

\paragraph{Toy model.}
The equivariant network contains $m$ scalar and $m$ vector channels, giving hidden dimension $4m$.
The input layer builds scalar features from each point's squared distance to the origin and vector features from its coordinates.
Three hidden layers mix scalar and vector channels separately, add learned combinations of squared vector lengths to the scalar features, and apply sigmoid gates to vectors and SiLU to scalars.
A final linear layer uses the scalar features to predict the largest eigenvalue.
The dense MLP projects flattened coordinates to dimension $4m$ and uses three SiLU layers followed by a final linear layer.

\paragraph{Interatomic potential models.}
Our NequIP implementation~\citep{batzner2022nequip,e3nn} uses two gated tensor product interaction layers with hidden irreps $32\times0e+32\times1o+32\times2e$, spherical harmonics up to $l_{\max}=2$, a $4.5\,\text{\AA}$ cutoff, and eight radial basis functions.
A radial network with hidden width 32 produces the tensor product weights.
Messages are summed over neighbors and mixed by equivariant linear layers, with SiLU activations for scalars and sigmoid gates for higher degrees.
GotenNet~\citep{aykent2025gotennet} uses 64 channels, two interaction layers, $l_{\max}=2$, four attention heads, a $5\,\text{\AA}$ cutoff, and 16 radial basis functions.
We enable edge updates and separate direction and tensor coefficients for each degree.
Its energy readout has a SiLU hidden layer of width 32.
Both models sum atomic energy predictions and obtain forces as negative energy gradients with respect to atomic positions.

\paragraph{QM9.}
We use the same GotenNet representation with twelve prediction heads.
Dipole moment and electronic spatial extent use the corresponding heads from the GotenNet implementation; the remaining heads predict atomic contributions that are summed over the molecule.
We center molecules, include atomic reference contributions where available, and compute normalization statistics from the training set.

\paragraph{Particle dynamics.}
We adapt NequIP to take charges, speeds, and velocity vectors as inputs, retaining two interaction layers with 32 channels per degree up to $l_{\max}=2$.
All distinct particle pairs are connected.
Edge features contain eight Gaussian radial basis functions of the distance and the product of the two particle charges.
A readout linear layer predicts a displacement, which is added to the input position.

\paragraph{Hyperparameters.}
We select learning rates and moment coefficients using mean validation error across seeds, and also search the initialization scale for the synthetic task.
In the synthetic ablation, moment coefficients selected with normalization are transferred to unnormalized Adam.
Test results are reported only for the selected configurations.
All runs use cosine learning rate decay to zero and no weight decay.

\begin{table*}[t]
	\centering
	\caption{Hyperparameters ranges. All runs use a cosine learning rate schedule, no warmup and no weight decay. Test error is evaluated at the checkpoint with the lowest validation error.}
	\label{tab:experimental-details}
	\setlength{\tabcolsep}{5pt}
	\renewcommand{\arraystretch}{1.1}
	\resizebox{\textwidth}{!}{%
		\begin{tabular}{lccccc}
			\toprule
			 &                                            & \multicolumn{2}{c}{rMD17 / MD22}        &                                         &                                                 \\
			\cmidrule(lr){3-4}
			 & Synthetic                                  & NequIP                                  & GotenNet                                & QM9                         & Particle dynamics \\
			\midrule
			Models
			 & \shortstack{e3nn,                                                                                                                                                                \\dense MLP} & NequIP & GotenNet & GotenNet & NequIP \\
			\cmidrule{1-6}
			Train/validation/test
			 & 8192/512/2048                              & 950/50/2000                             & 950/50/2000                             & 110{,}000/10{,}000/10{,}831 & 3000/2000/2000    \\
			\cmidrule{1-6}
			Training steps
			 & 1000                                       & 5000                                    & 5000                                    & 200{,}000                   & 30{,}000          \\
			\cmidrule{1-6}
			Batch size
			 & Full batch                                 & 32 / 16                                 & 32 / 16                                 & 32                          & 100               \\
			\cmidrule{1-6}
			Evaluation interval
			 & 100 steps                                  & 250 steps                               & 250 steps                               & 2500 steps                  & 250 steps         \\
			\cmidrule{1-6}
			Seeds
			 & 8                                          & 3                                       & 3                                       & 3                           & 3                 \\
			\cmidrule{1-6}
			Loss
			 & MSE                                        & $\mathrm{MSE}(E) + 10\,\mathrm{MSE}(F)$ & $\mathrm{MSE}(E) + 10\,\mathrm{MSE}(F)$ & MSE                         & MSE               \\
			\midrule
			Adam $(\beta_1, \beta_2)$
			 & \shortstack{$\beta_1 \in [0, 0.99]$,                                                                                                                                             \\$\beta_2 \in [0.9, 0.9999]$}
			 & \shortstack{$(0.9, 0.999)$,                                                                                                                                                      \\$(0.98, 0.99)$}
			 & \shortstack{$(0.9, 0.999)$,                                                                                                                                                      \\$(0.98, 0.99)$}
			 & \shortstack{$(0.9, 0.999)$, $(0.9, 0.99)$,                                                                                                                                       \\$(0.98, 0.99)$, $(0.99, 0.99)$}
			 & \shortstack{$(0.9, 0.999)$,                                                                                                                                                      \\$(0.98, 0.99)$} \\
			\cmidrule{1-6}
			Muon momentum
			 & $0.95$                                     & $0.95$, $0.99$                          & $0.95$, $0.99$                          & -                           & -                 \\
			\midrule
			Adam learning rate
			 & $0.0001$-$0.1$                             & $0.003$-$0.3$                           & $0.001$-$0.03$                          & $0.0003$-$0.01$             & $0.0003$-$0.003$  \\
			\cmidrule{1-6}
			Spectral learning rate
			 & $0.0001$-$0.1$                             & $0.03$-$1$                              & $0.003$-$0.3$                           & $0.003$-$0.1$               & $0.003$-$0.03$    \\
			\cmidrule{1-6}
			RMS learning rate
			 & -                                          & $0.001$-$0.03$                          & $0.0001$-$0.01$                         & $0.0001$-$0.003$            & $0.0001$-$0.001$  \\
			\cmidrule{1-6}
			Muon learning rate
			 & $0.0001$-$0.1$                             & $0.01$-$0.3$                            & $0.001$-$0.1$                           & -                           & -                 \\
			\midrule
			SGD learning rate
			 & -                                          & $0.0001$-$0.01$                         & -                                       & -                           & -                 \\
			\cmidrule{1-6}
			SGD + spectral learning rate
			 & -                                          & $0.03$-$1$                              & -                                       & -                           & -                 \\
			\cmidrule{1-6}
			SGD + RMS learning rate
			 & -                                          & $0.001$-$0.01$                          & -                                       & -                           & -                 \\
			\cmidrule{1-6}
			SGD momentum
			 & -                                          & $0.9$, $0.99$                           & -                                       & -                           & -                 \\
			\cmidrule{1-6}
			SGD + spectral, RMS momentum
			 & -                                          & $0.95$, $0.99$                          & -                                       & -                           & -                 \\
			\midrule
			Initialization scale
			 & $0.03$, $0.1$, $0.3$, $1$                  & Default                                 & Default                                 & Default                     & Default           \\
			\bottomrule
		\end{tabular}%
	}
\end{table*}

Table~\ref{tab:selected-hyperparameters} lists the variants used for the main results.
All selected synthetic variants use initialization scale $0.3$.
For molecular force prediction, learning rates and moment coefficients are selected separately for each model and dataset.

\begin{table*}[t]
	\centering
	\caption{Selected hyperparameters.}
	\label{tab:selected-hyperparameters}
	\small
	\setlength{\tabcolsep}{6pt}
	\renewcommand{\arraystretch}{1.1}
	\begin{tabular}{llcc}
		\toprule
		Dataset & Method             & $(\beta_1, \beta_2)$ & Learning rate      \\
		\midrule
		\multirow{6}{*}{Synthetic}
		        & Adam               & $(0.9, 0.999)$       & $0.001$, $0.00032$ \\
		        & Spectral           & $(0.9, 0.999)$       & $0.01$             \\
		        & + transferred mom. & $(0.5, 0.99)$        & $0.001$, $0.00032$ \\
		        & + both             & $(0.5, 0.99)$        & $0.01$             \\
		        & Muon               & $0.95$               & $0.0032$           \\
		\midrule
		\multirow{4}{*}{\shortstack[l]{NequIP,                                   \\rMD17 / MD22}}
		        & Adam               & $(0.98, 0.99)$       & $0.03$, $0.1$      \\
		        & Spectral           & $(0.98, 0.99)$       & $0.1$              \\
		        & RMS                & $(0.98, 0.99)$       & $0.003$            \\
		        & Muon               & $0.99$               & $0.03$, $0.1$      \\
		\midrule
		\multirow{4}{*}{\shortstack[l]{GotenNet,                                 \\rMD17 / MD22}}
		        & Adam               & $(0.98, 0.99)$       & $0.003$, $0.01$    \\
		        & Spectral           & $(0.98, 0.99)$       & $0.01$             \\
		        & RMS                & $(0.98, 0.99)$       & $0.0003$           \\
		        & Muon               & $0.99$               & $0.003$, $0.01$    \\
		\midrule
		\multirow{4}{*}{\shortstack[l]{GotenNet,                                 \\QM9}}
		        & Adam               & $(0.98, 0.99)$       & $0.00095$          \\
		        & Spectral           & $(0.98, 0.99)$       & $0.0032$           \\
		        & RMS                & $(0.9, 0.99)$        & $0.0003$           \\
		\midrule
		\multirow{4}{*}{\shortstack[l]{NequIP,                                   \\particle dynamics}}
		        & Adam               & $(0.9, 0.999)$       & $0.003$            \\
		        & Spectral           & $(0.9, 0.999)$       & $0.03$             \\
		        & RMS                & $(0.9, 0.999)$       & $0.001$            \\
		\midrule
		\multirow{3}{*}{\shortstack[l]{NequIP,                                   \\aspirin, SGD}}
		        & SGD                & $0.99$               & $0.001$            \\
		        & Spectral           & $0.99$               & $0.1$              \\
		        & RMS                & $0.99$               & $0.003$            \\
		\bottomrule
	\end{tabular}
\end{table*}

\begin{table}[h]
	\centering
	\caption{Test force MAE (kcal/mol/\AA, $\downarrow$) of Muon with the exact polar update (SVD) and with five Newton-Schulz iterations, for NequIP on rMD17 aspirin and ethanol. Both use momentum 0.99 and learning rate 0.1, the Newton-Schulz selection of Table~\ref{tab:force-mae-nequip}. Test error is read at the checkpoint with the lowest validation error (mean $\pm$ standard deviation over three seeds).}
	\label{tab:muon-implementation}
	\begin{tabular}{lcc}
		\toprule
		Muon update   & Aspirin           & Ethanol           \\
		\midrule
		Exact SVD     & $0.684 \pm 0.012$ & $0.268 \pm 0.016$ \\
		Newton-Schulz & $0.679 \pm 0.005$ & $0.262 \pm 0.001$ \\
		\bottomrule
	\end{tabular}
\end{table}

Muon uses Nesterov momentum with the coefficients listed in Table~\ref{tab:selected-hyperparameters}.
On the synthetic task and in Figure~\ref{fig:block-step-normalization}, Muon computes the exact polar update by SVD. %
In all other experiments, it approximates the polar update with five Newton-Schulz iterations in bfloat16.
Table~\ref{tab:muon-implementation} compares the two on rMD17 aspirin and ethanol with NequIP, with the difference between them being negligible.

\section{Additional Empirical Results}

\begin{figure}[t]
	\centering
	\includegraphics[width=0.5\linewidth]{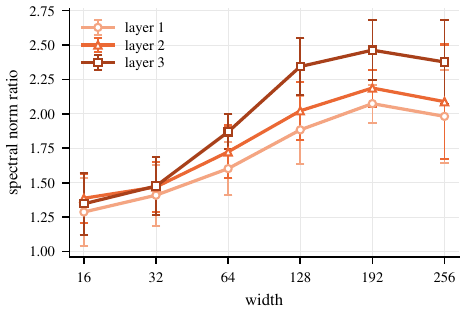}
	\caption{Ratio between the spectral norms of the updates of the scalar and vector irrep blocks, $W_0$ and $W_1$, in each hidden layer of the equivariant model on the synthetic task, at Adam's first update. Setting as in Figure~\ref{fig:toy-step-magnitudes}; mean and standard deviation over eight seeds.}
	\label{fig:toy-within-layer}
\end{figure}

Figure~\ref{fig:toy-within-layer} splits the ratio of Figure~\ref{fig:toy-step-magnitudes} by hidden layer.
In every layer, the ratio between the two irrep blocks grows with width.

Table~\ref{tab:force-mae-gotennet} reports the complete GotenNet results across rMD17 and two MD22 datasets.
Both normalizations improve Adam on every dataset, but their ranking reverses compared with NequIP in Table~\ref{tab:force-mae-nequip}.
Spectral normalization generally performs better than RMS for GotenNet, whereas RMS consistently performs better for NequIP.

\begin{table}[t]
	\centering
	\caption{Force MAE (kcal/mol/\AA; mean $\pm$ standard deviation over three seeds; lower is better) for GotenNet on rMD17 and MD22. Each optimizer uses its own validation-selected learning rate and moment coefficients; test error is read at the checkpoint with the lowest validation error. Bold marks the best result per dataset, underline the second best.}
	\label{tab:force-mae-gotennet}
	\small
	\setlength{\tabcolsep}{5pt}
	\begin{tabular}{clcccc}
		\toprule
		 & Dataset        & Adam              & Spectral                      & RMS                           & Muon                       \\
		\midrule
		\multirow{10}{*}{\rotatebox[origin=c]{90}{\scriptsize rMD17}}
		 & ethanol        & $0.194 \pm 0.001$ & $\underline{0.152 \pm 0.005}$ & $0.165 \pm 0.004$             & $\mathbf{0.092 \pm 0.001}$ \\
		 & malonaldehyde  & $0.289 \pm 0.006$ & $\underline{0.232 \pm 0.006}$ & $0.237 \pm 0.003$             & $\mathbf{0.150 \pm 0.004}$ \\
		 & benzene        & $0.052 \pm 0.004$ & $\underline{0.038 \pm 0.002}$ & $\underline{0.038 \pm 0.003}$ & $\mathbf{0.019 \pm 0.001}$ \\
		 & uracil         & $0.205 \pm 0.014$ & $\underline{0.172 \pm 0.004}$ & $0.175 \pm 0.007$             & $\mathbf{0.109 \pm 0.003}$ \\
		 & toluene        & $0.167 \pm 0.012$ & $\underline{0.134 \pm 0.005}$ & $0.135 \pm 0.004$             & $\mathbf{0.084 \pm 0.003}$ \\
		 & salicylic acid & $0.311 \pm 0.008$ & $\underline{0.247 \pm 0.001}$ & $0.251 \pm 0.004$             & $\mathbf{0.166 \pm 0.002}$ \\
		 & naphthalene    & $0.209 \pm 0.006$ & $\underline{0.141 \pm 0.007}$ & $0.144 \pm 0.003$             & $\mathbf{0.092 \pm 0.005}$ \\
		 & paracetamol    & $0.388 \pm 0.011$ & $\underline{0.295 \pm 0.003}$ & $0.304 \pm 0.006$             & $\mathbf{0.210 \pm 0.002}$ \\
		 & aspirin        & $0.432 \pm 0.025$ & $\underline{0.365 \pm 0.012}$ & $0.376 \pm 0.002$             & $\mathbf{0.280 \pm 0.004}$ \\
		 & azobenzene     & $0.326 \pm 0.023$ & $\underline{0.232 \pm 0.005}$ & $0.234 \pm 0.008$             & $\mathbf{0.154 \pm 0.003}$ \\
		\cmidrule{2-6}
		\multirow{2}{*}{\rotatebox[origin=c]{90}{\scriptsize MD22}}
		 & Ac-Ala3-NHMe   & $0.753 \pm 0.018$ & $0.598 \pm 0.020$             & $\underline{0.596 \pm 0.012}$ & $\mathbf{0.398 \pm 0.005}$ \\
		 & stachyose      & $1.589 \pm 0.033$ & $\underline{1.245 \pm 0.012}$ & $1.266 \pm 0.025$             & $\mathbf{0.960 \pm 0.013}$ \\
		\bottomrule
	\end{tabular}
\end{table}

\end{document}